\documentclass{article}
\usepackage{spconf,amsmath,graphicx,amssymb,amsthm,comment,bm,cite,url}
\usepackage{array}
\allowdisplaybreaks

\newcommand{\refeq}[1]{(\ref{eq:#1})}
\newcommand{\refeqs}[2]{(\ref{eq:#1}) and (\ref{eq:#2})}

\newcommand{\refsubsec}[1]{\ref{subsec:#1}}
\newcommand{\reffig}[1]{Fig. \ref{fig:#1}}
\newcommand{\reffigs}[2]{Figs. \ref{fig:#1} and \ref{fig:#2}}

\def\Vec#1{\boldsymbol{\mathbf{#1}}}

\def\x{\Vec{x}}
\def\s{\Vec{s}}
\def\e{\Vec{e}}
\def\v{\Vec{v}}
\def\y{\Vec{y}}

\def\xH{\Vec{x}^{\mathsf H}}

\def\w{\Vec{w}}
\def\wH{\Vec{w}^{\mathsf H}}
\def\W{\Vec{W}}
\def\V{\Vec{V}}
\def\WH{\Vec{W}^{\mathsf H}}

\def\S{\Vec{S}}
\def\sh{\boldsymbol{\mathsf h}}
\def\sW{\boldsymbol{\mathsf W}}
\def\sV{\boldsymbol{\mathsf V}}
\def\sb{\boldsymbol{\mathsf b}}
\def\sd{\boldsymbol{\mathsf d}}
\def\BSigma{\Vec{\Sigma}}
\def\zero{\Vec{0}}
\def\encdis{q_{\phi}}
\def\decdis{p_{\theta}}
\def\CX{{\mathcal X}}
\def\CS{{\mathcal S}}
\def\CW{{\mathcal W}}
\def\CV{{\mathcal V}}
\def\CB{{\mathcal B}}
\def\CH{{\mathcal H}}
\def\CG{{\mathcal G}}
\def\CN{{\mathcal N}}

\def\z{\Vec{z}}

\def\0{{\mathbf 0}}
\def\I{{\mathbf I}}
\def\vmu{{\boldsymbol \mu}}
\def\vsigma{{\boldsymbol \sigma}}
\def\vepsilon{{\boldsymbol \epsilon}}

\makeatletter
\newcommand{\thickhline}{%
    \noalign {\ifnum 0=`}\fi \hrule height 1pt
    \futurelet \reserved@a \@xhline
}
\newcolumntype{"}{@{\hskip\tabcolsep\vrule width 1pt\hskip\tabcolsep}}
\makeatother

\title{
Semi-Blind Source Separation with\\
Multichannel Variational Autoencoder
}
\name{
Hirokazu Kameoka$^1$, Li Li$^2$, Shota Inoue$^2$, Shoji Makino$^2$
}
\address{
$^1$ NTT Communication Science Laboratories, NTT Corporation, Japan\\
$^2$ University of Tsukuba, Japan
}
\begin{document}
%
\maketitle
\begin{abstract}
This paper proposes a multichannel source separation technique called the multichannel variational autoencoder (MVAE) method, which uses a conditional VAE (CVAE) to model and estimate the power spectrograms of the sources in a mixture. 
By training the CVAE using the spectrograms of training examples with source-class labels, we can use the trained decoder distribution as a universal generative model capable of generating spectrograms conditioned on a specified class label.
By treating the latent space variables and the class label as the unknown parameters of this generative model, we can develop a convergence-guaranteed semi-blind source separation algorithm that consists of iteratively estimating the power spectrograms of the underlying sources as well as the separation matrices.  
In experimental evaluations, our MVAE produced better separation performance than a baseline method.
\end{abstract}
\section{Introduction}
\label{sec:intro}

Blind source separation (BSS) is a technique for separating out individual source signals from microphone array inputs when the transfer characteristics between the sources and microphones are unknown. 
The frequency-domain BSS approach provides the flexibility of allowing
us to utilize various models for the time-frequency representations
of source signals and/or array responses. For example, independent
vector analysis (IVA) \cite{Kim2006,Hiroe2006} allows us to efficiently solve
frequency-wise source separation and permutation alignment in a
joint manner by assuming that the magnitudes of the frequency components
originating from the same source tend to vary coherently
over time.

With a different approach, multichannel extensions of nonnegative matrix factorization (NMF) have attracted a lot of attention
in recent years \cite{Ozerov2010,Kameoka2010,Sawada2013,Kitamura2016,Kitamura2017}. 
NMF was originally applied to music transcription and 
monaural source separation tasks \cite{Smaragdis2003,Fevotte2009}. 
The idea is to approximate the power (or magnitude) spectrogram of a mixture signal, interpreted as a non-negative matrix, as the product of two non-negative matrices.
This amounts to assuming that the power spectrum of a mixture signal observed at each time frame can be approximated by the linear
sum of a limited number of basis spectra scaled by time-varying amplitudes. 
Multichannel NMF (MNMF) is an extension of this approach to a multichannel case to allow the use of spatial information as an additional clue to separation. It can also be viewed as an extension of frequency-domain BSS that allows the use of spectral templates as a clue for 
jointly solving frequency-wise source separation and permutation alignment.

The original MNMF \cite{Ozerov2010} was formulated under a general problem setting where sources can outnumber microphones and a determined version of MNMF was subsequently proposed in \cite{Kameoka2010}. 
While the determined version is applicable only to determined cases, it allows the implementation of a significantly faster algorithm than the general version. 
The determined MNMF framework was later called ``independent low-rank matrix analysis (ILRMA)'' \cite{Kitamura2017}.
In \cite{Kitamura2016}, the theoretical relation of MNMF to IVA was discussed, which has naturally allowed for the incorporation
of the fast update rule of the separation matrix developed for IVA, called ``iterative projection (IP)'' \cite{Ono2011}, into the parameter optimization process in ILRMA. It has been shown that this has contributed not only to further accelerating the entire optimization process but also to improving the separation performance. While ILRMA is notable in that the optimization algorithm is guaranteed to converge, it can fail to work for sources with spectrograms that do not comply with the NMF model. 

As an alternative to the NMF model, some attempts have recently been made to 
use deep neural networks (DNNs) for modeling the spectrograms of sources for multichannel source separation \cite{Nugraha2016,Kitamura2018}.
The idea is to replace the process for estimating the power spectra of source signals
in a source separation algorithm with the forward computations of pretrained DNNs. 
This can be viewed as a process of refining the estimates of the power spectra of the source signals 
at each iteration of the algorithm. 
While this approach is particularly appealing in that it 
can take advantage of the strong representation power of DNNs for estimating 
the power spectra of source signals, one weakness is that 
the convergence of an algorithm devised in this way 
will not be guaranteed. 

To address the drawbacks of the methods mentioned above, 
this paper proposes a multichannel source separation method 
using variational autoencoders (VAEs) \cite{Kingma2014a,Kingma2014b} for source spectrogram modeling.  
We call our approach the ``multichannel VAE (MVAE)'' method.

\section{Problem formulation}

We consider a situation where $I$ source signals are captured by $I$ microphones.
Let $x_i(f,n)$ and 
$s_j(f,n)$ be the short-time Fourier transform (STFT) coefficients 
of the signal observed at the $i$-th microphone and the $j$-th source signal, where $f$ and $n$ are
the frequency and time indices, respectively. We denote 
the vectors containing $x_1(f,n),\ldots,x_I(f,n)$ and 
$s_1(f,n),\ldots,s_I(f,n)$ by
\begin{align}
\Vec{x}(f,n) &= [x_1(f,n),\ldots,x_I(f,n)]^{\mathsf T}\in\mathbb{C}^{I},\\
\Vec{s}(f,n) &= [s_1(f,n),\ldots,s_I(f,n)]^{\mathsf T}\in\mathbb{C}^{I},
\end{align}
where $(\cdot)^{\mathsf T}$ denotes transpose.
Now, we use a separation system of the form
\begin{align}
\Vec{s}(f,n) &= \Vec{W}^{\mathsf H}(f) \Vec{x}(f,n),
\label{eq:sepsystem}\\
\W(f) &= [\w_1(f),\ldots,\w_I(f)],
\end{align}
to describe the relationship between $\Vec{x}(f,n)$ and $\Vec{s}(f,n)$
where $\Vec{W}^{\mathsf H}(f)$ is usually called the separation matrix.
$(\cdot)^{\mathsf H}$ denotes Hermitian transpose.
The aim of BSS methods is to estimate $\Vec{W}^{\mathsf H}(f)$ solely 
from the observation $\Vec{x}(f,n)$.

Let us now assume that 
$s_j(f,n)$
independently follows a zero-mean complex Gaussian distribution with
variance $v_j(f,n)=\mathbb{E}[|s_j(f,n)|^2]$
\begin{align}
s_j(f,n) \sim \mathcal{N}_{\mathbb{C}}(s_j(f,n)|0,v_j(f,n)).
\label{eq:LGM1}
\end{align}
We call \refeq{LGM1} the local Gaussian model (LGM).
When $s_j(f,n)$ and $s_{j'}(f,n)$ $(j\neq j')$ are independent, 
$\s(f,n)$ follows
\begin{align}
\s(f,n) \sim \mathcal{N}_{\mathbb{C}}(\s(f,n)|\zero, \V(f,n)),
\label{eq:LGM2}
\end{align}
where $\V(f,n)$ is a diagonal matrix with diagonal entries $v_1(f,n),\ldots,v_I(f,n)$.
From \refeqs{sepsystem}{LGM1}, we can show that $\x(f,n)$ follows
\begin{align}
\x(f,n)\sim \mathcal{N}_{\mathbb{C}}(\x(f,n)|\zero, (\WH(f))^{-1}\V(f,n)\W(f)^{-1}).
\end{align}
Hence, the log-likelihood of the separation matrices $\CW=\{\W(f)\}_f$ 
given the observed mixture signals $\CX=\{\x(f,n)\}_{f,n}$ 
is given by
\begin{align}
&\log p(\CX|\CW,\CV) \mathop{=}^c 
2N \sum_{f} \log|\det \WH(f)| 
\nonumber\\
&~~~~~~~~~~~- 
\sum_{f,n}\sum_j 
\bigg(\log v_j(f,n) + \frac{|\wH_j(f)\x(f,n)|^2}{v_j(f,n)}\bigg),
\label{eq:loglikelihood}
\end{align}
where $\mathop{=}^c$ denotes equality up to constant terms.
If we individually treat $v_j(f,n)$ as a free parameter, 
all the variables in \refeq{loglikelihood} will be indexed by frequency $f$. 
The optimization problem will thus be split into frequency-wise source separation problems.
Under this problem setting, the permutation of the separated components in each frequency cannot be uniquely determined and so permutation alignment must be performed after $\CW$ has been obtained. 
However, it is preferable to solve permutation alignment
and source separation jointly since the clues used for permutation alignment
can also be helpful for source separation.
If there is a certain assumption, constraint or structure that we can incorporate into 
$v_j(f,n)$, it can help eliminate the permutation ambiguity during the estimation of $\CW$. 
One such example is the NMF model, which expresses $v_j(f,n)$ as the linear sum of 
spectral templates $b_{j,1}(f),\ldots,b_{j,K_j}(f)\ge 0$ scaled by time-varying magnitudes 
$h_{j,1}(n),\ldots,h_{j,K_j}(n)\ge 0$:
\begin{align}
v_j(f,n) = \sum_{k=1}^{K_j} b_{j,k}(f) h_{j,k}(n).
\label{eq:nmf_model}
\end{align}
ILRMA is a BSS framework that incorporates this model into the log-likelihood \refeq{loglikelihood} \cite{Kameoka2010,Kitamura2016,Kitamura2017}.
Here, in a particular case where $K_j=1$ and $b_{j,k}(f)=1$ for all $j$ in \refeq{nmf_model}, 
which means each source has only one flat-shaped spectral template, assuming 
$s_j(0,n),\ldots,s_j(F,n)$ independently follow \refeq{LGM1} is equivalent to assuming the norm $r_j(n) = \sqrt{\sum_f|s_j(f,n)|^2}$ follows a Gaussian distribution with time-varying variance $h_j(n)$.
This is analogous to the assumption employed by IVA where the norm 
$r_j(n)$
is assumed to follow a supergaussian distribution.
\cite{Kitamura2016} showed that ILRMA can significantly outperform 
IVA in terms of source separation ability. This fact implies that
within the LGM-based BSS framework, the stronger the representation power of a power spectrogram model becomes, the better the source separation performance we can expect to obtain.

\section{Related work}

\subsection{ILRMA}

The optimization algorithm of 
ILRMA
consists of iteratively updating 
$\CW$, $\CB=\{b_{j,k}(f)\}_{j,k,f}$ and $\CH=\{h_{j,k}(n)\}_{j,k,n}$ 
so that \refeq{loglikelihood} is guaranteed to be non-decreasing at each iteration \cite{Kameoka2010,Kitamura2016,Kitamura2017}.
To update $\CW$,
we can use the natural gradient method or IP. 
The IP-based update rule for $\CW$ \cite{Ono2011} is given as
\begin{align}
\w_j(f) &\leftarrow (\WH(f)\BSigma_j(f))^{-1} \e_j,
\label{eq:w_update1}\\
\w_j(f) &\leftarrow
\frac{\w_j(f)}{\wH_j(f)\BSigma_j(f)\w_j(f)},
\label{eq:w_update2}
\end{align}
where $\BSigma_j(f)=\frac{1}{N}\sum_n \x(f,n)\xH(f,n)/v_j(f,n)$
and $\e_j$ denotes the $j$-th column of the $I \times I$ identity matrix. 
To update $\CB$ and $\CH$, we can employ the expectation-maximization (EM) 
algorithm or the majorization-minimization (MM) algorithm.
The MM-based update rules for $\CB$ and $\CH$ can be derived \cite{Kameoka2006,Nakano2010,Fevotte2011} as
\begin{align}
b_{j,k}(f) &\leftarrow 
b_{j,k}(f)
\sqrt{
\frac{
\sum_{n} |y_j(f,n)|^2 h_{j,k}(n)/v_{j}^2(f,n)
}{
\sum_{n} h_{j,k}(n)/v_{j}(f,n)
}
},\!\!
\label{eq:b_update}
\\
h_{j,k}(n) &\leftarrow
h_{j,k}(n)
\sqrt{
\frac{
\sum_{f} |y_j(f,n)|^2 b_{j,k}(f)/v_{j}^2(f,n)
}{
\sum_{f} b_{j,k}(f)/v_{j}(f,n)
}
},\!\!
\label{eq:h_update}
\end{align}
where $y_j(f,n) = \wH_j(f)\x(f,n)$.

\begin{figure}[t!]
\centering
\begin{minipage}{.8\linewidth}
  \centerline{\includegraphics[width=.98\linewidth]{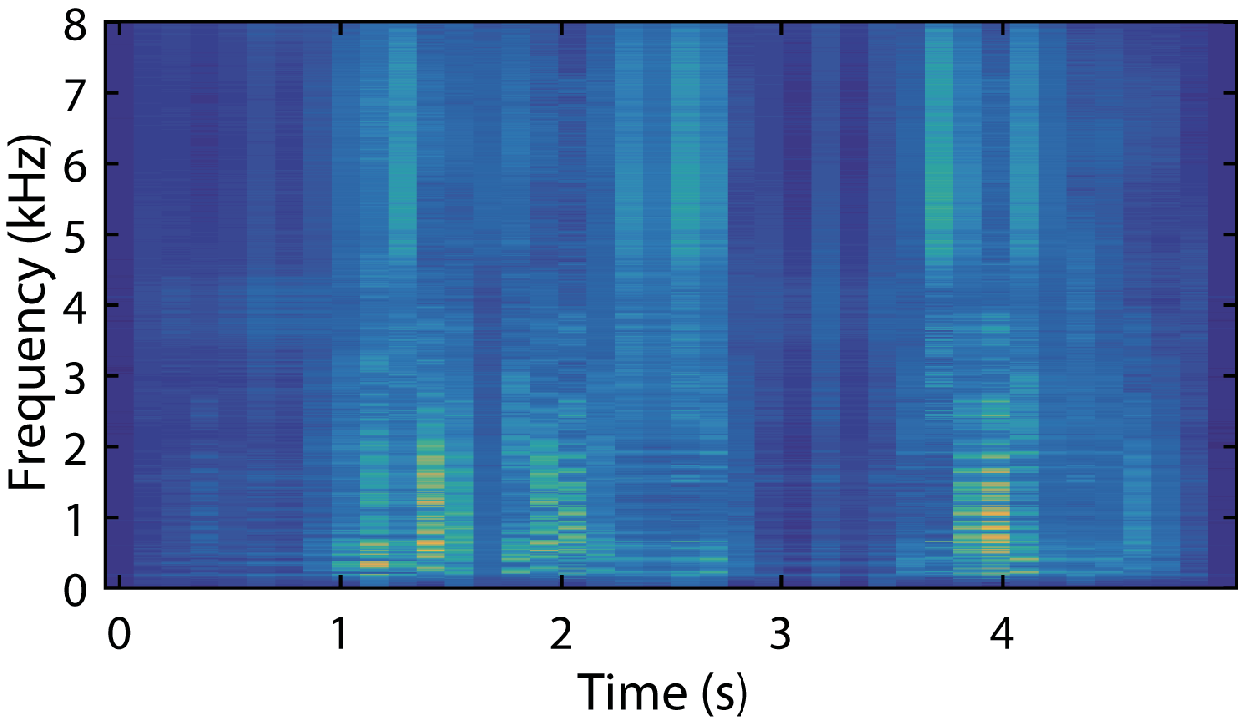}}
  \centerline{\includegraphics[width=.98\linewidth]{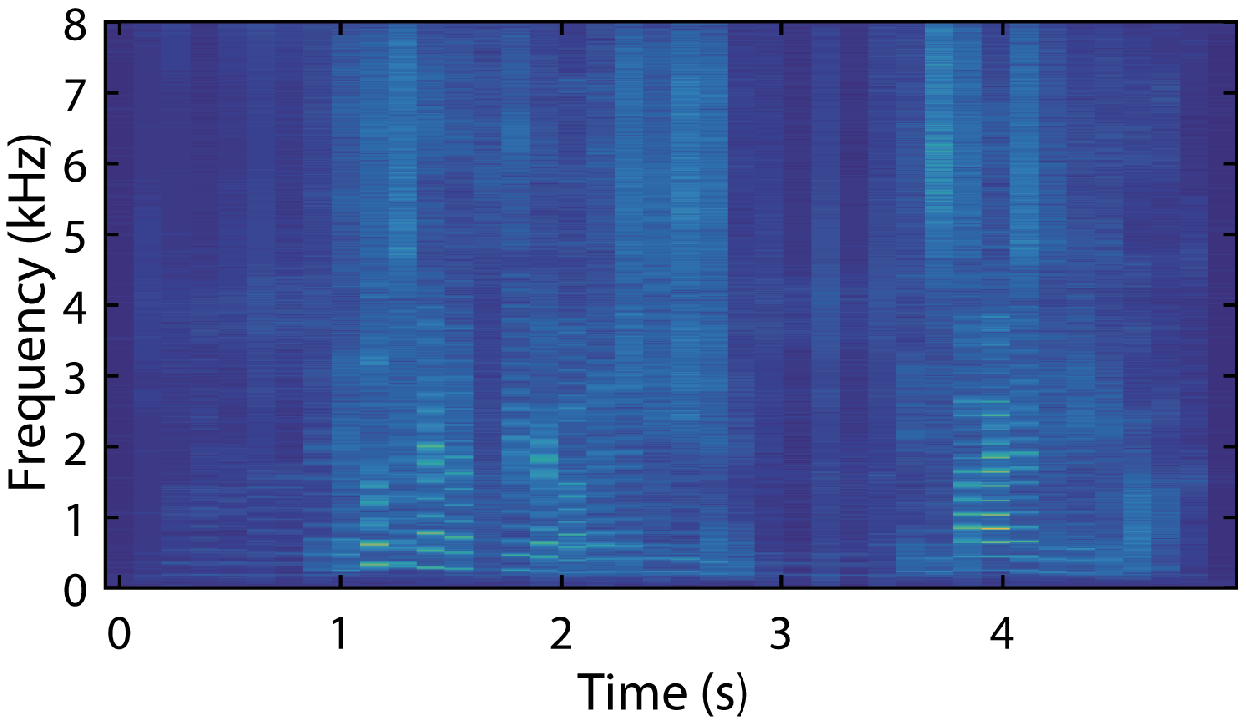}}
  \vspace{-1ex}
  \caption{Example of the NMF model (top) fitted to a speech spectrogram (bottom).}
\label{fig:nmfmodel}
\end{minipage}
\end{figure}

ILRMA is notable in that the optimization algorithm 
is guaranteed to converge to a stationary point of \refeq{loglikelihood}
and is shown experimentally to converge quickly.
However, one limitation is that since
$v_j(f,n)$ is restricted to \refeq{nmf_model}, it can fail to work 
for sources with spectrograms that do not actually follow \refeq{nmf_model}.
\reffig{nmfmodel} shows an example of the NMF model optimally 
fitted to a speech spectrogram. As can be seen from this example, there is 
still plenty of room for improvement in the model design.

\subsection{DNN approach}

As an alternative to the NMF model, some attempts have recently been made to 
combine 
deep neural networks (DNNs) 
with
the LGM-based multichannel source separation framework
\cite{Nugraha2016,Kitamura2018}.
\cite{Nugraha2016,Kitamura2018} propose algorithms
where $v_j(f,n)$ is updated at each iteration
to the output of pretrained DNNs
\begin{align}
\tilde{\v}_j(n)\leftarrow {\rm DNN}(\tilde{\y}_j(n);\theta_j)~~(n=1,\ldots,N).
\label{eq:v_update_dnn}
\end{align}
Here, 
${\rm DNN}(\cdot;\theta_j)$  indicates the output of the pretrained DNN,
$\theta_j$ is the set of NN parameters,  
$\tilde{\y}_j(n)=
\{|y_j(f,n\pm n')|\}_{f,n'}$
denotes the magnitude spectra of the estimate of 
the $j$-th separated signal around the $n$-th time frame and 
$\tilde{\v}_j(n)=\{\sqrt{v_j(f,n)}\}_f$. 
With this approach, multiple DNNs are trained, and 
the $j$-th DNN is trained so that it produces only
spectra related to source $j$ in noisy input spectra.
\refeq{v_update_dnn} can thus be seen as a process of refining the magnitude spectra of
the separated signals 
according to the training examples of the known sources.

While this approach is noteworthy in that 
it can exploit the benefits of the representation power of DNNs for source power spectrum modeling,
one drawback is that the devised iterative algorithm is not guaranteed to converge 
to a stationary point of the log-likelihood 
since updating $v_j(f,n)$ in this way 
does not guarantee an increase in the log-likelihood. 

\subsection{Source separation using deep generative models}

It is worth noting that there have been some attempts to apply
deep generative models including 
VAEs \cite{Kingma2014a,Kingma2014b}
and generative adversarial networks (GANs) \cite{Goodfellow2014}
to monaural speech enhancement and source separation \cite{Bando2018,Subakan2018}. 
However, to the best of our knowledge, their applications to
multichannel source separation has yet to be proposed.

\section{Proposed method}

To address the limitations and drawbacks of the conventional methods, 
this paper proposes a multichannel source separation method using
VAEs for source spectrogram modeling. 
We briefly review the idea behind the VAEs in \refsubsec{vae}
and present the proposed source separation algorithm in \refsubsec{propalgo}, which we call 
the multichannel VAE (MVAE).

\subsection{Variational autoencoder (VAE)}
\label{subsec:vae}

\begin{figure*}[t!]
\centering
  \begin{minipage}{.7\linewidth}
  \centerline{\includegraphics[width=.98\linewidth]{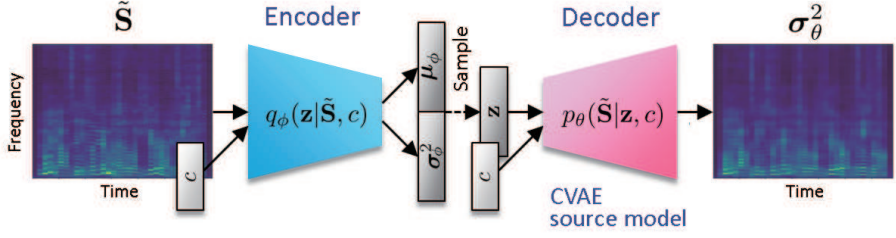}}
  \vspace{-3ex}
  \caption{Illustration of the present CVAE.}
  \label{fig:cvae_training}
  \end{minipage}
\end{figure*}

VAEs \cite{Kingma2014a,Kingma2014b} are stochastic neural network models consisting of encoder and decoder networks.
The encoder network generates a set of parameters for the conditional distribution $\encdis(\z|\s)$ of a latent space variable $\z$ given input data $\s$ whereas 
the decoder network generates a set of parameters for the conditional distribution $\decdis(\s|\z)$ of the data $\s$ given the latent space variable $\z$. 
Given a training dataset $\CS = \{ \s_m \}_{m=1}^{M}$, 
VAEs learn the parameters of
the entire network so that the encoder distribution $\encdis(\z|\s)$ becomes consistent with the posterior $\decdis(\z|\s)\propto \decdis(\s|\z)p(\z)$. 
By using Jensen's inequality, the log marginal distribution of the data $\s$ can be lower-bounded by
\begin{align}
\log \decdis(\s) &
= \log \int \encdis(\z|\s) \frac{\decdis(\s|\z)p(\z)}{\encdis(\z|\s)} d\z 
\nonumber\\
&\ge
\int \encdis(\z|\s) \log \frac{\decdis(\s|\z)p(\z)}{\encdis(\z|\s)} d\z
\label{eq:lowerbound}\\
&=
\mathbb{E}_{\z\sim \encdis(\z|\s)}[\log \decdis(\s|\z)] 
- {\rm KL}[ \encdis(\z|\s) \| p(\z) ],
\nonumber
\end{align}
where the difference between the left- and right-hand sides of this inequality 
is equal to the Kullback-Leibler divergence ${\rm KL}[\encdis(\z|\s) \| \decdis(\z|\s)]$,
which is minimized when
\begin{align}
\encdis(\z|\s) = \decdis(\z|\s).
\end{align}
This means we can make $\encdis(\z|\s)$ and $\decdis(\z|\s)\propto \decdis(\s|\z)p(\z)$ 
consistent by maximizing the lower bound of \refeq{lowerbound}.
One typical way of modeling $\encdis(\z|\s)$, $\decdis(\s|\z)$ and $p(\z)$ is to assume
Gaussian distributions
\begin{align}
\encdis(\z|\s) &= \mathcal{N}(\z|\vmu_{\phi}(\s), {\rm diag}(\vsigma_{\phi}^2(\s))),
\label{eq:q(z|s)}
\\
\decdis(\s|\z) &= \mathcal{N}(\s|\vmu_{\theta}(\z), {\rm diag}(\vsigma_{\theta}^2(\z))),
\label{eq:p(s|z)}
\\
p(\z) &= \mathcal{N}(\z|{\bf 0},{\bf I}),
\label{eq:p(z)}
\end{align}
where $\vmu_{\phi}(\s)$ and $\vsigma_{\phi}^2(\s)$ are the outputs of an encoder network with parameter $\phi$,
and $\vmu_{\theta}(\z)$ and $\vsigma_{\theta}^2(\z)$ are the outputs of a decoder network with parameter $\theta$.
The first term of the lower bound can be interpreted as an autoencoder reconstruction error 
since it can be written as
\begin{align}
&\mathbb{E}_{\z\sim q(\z|\s)}[\log p(\s|\z)] \nonumber\\
=&
\mathbb{E}_{\vepsilon \sim \mathcal{N}(\vepsilon|\0,\I)}
\bigg[
-\frac{1}{2}
\sum_i
\log 2\pi 
[
\vsigma_{\theta}^2(
\vmu_{\phi}(\s) +
\vsigma_{\phi}(\s) \odot 
\vepsilon 
)
]_n
\nonumber\\
&~~~~~~~~~~
- 
\sum_n
\frac{
( 
s_n - 
[\vmu_{\theta}
(
\vmu_{\phi}(\s) +
\vsigma_{\phi}(\s) \odot
\vepsilon 
)
]_n
)^2
}{2
[\vsigma_{\theta}^2(
\vmu_{\phi}(\s) + 
\vsigma_{\phi}(\s) \odot
\vepsilon 
)]_n
}
\bigg],
\end{align}
which reduces to a negative weighted squared error between 
$\s$ and $\vmu_{\theta}(\vmu_{\phi}(\s))$  
if we exclude all the stochastic terms related to $\vepsilon$.
Here, we have used a reparameterization $\z = \vmu_{\phi}(\s) + \vsigma_{\phi}(\s) \odot \vepsilon$ with 
$\vepsilon \sim \mathcal{N}(\vepsilon|\0,\I)$ where $\odot$ indicates the element-wise product
and $[\cdot]_n$ denotes the $n$-th element of a vector. 
On the other hand, the second term is given as the negative KL divergence 
between $\encdis(\z|\s)$ and $p(\z)=\mathcal{N}(\z|\0,\I)$. This term can be interpreted as a regularization term that 
forces each element of the encoder output to be independent and normally distributed.

Conditional VAEs (CVAEs) \cite{Kingma2014b} are an extended version of VAEs where the only difference is that
the encoder and decoder networks can take an auxiliary variable 
$c$ as an additional input. With CVAEs, \refeqs{q(z|s)}{p(s|z)} are replaced with
\begin{align}
\encdis(\z|\s,c) &= \mathcal{N}(\z|\vmu_{\phi}(\s,c), {\rm diag}(\vsigma_{\phi}^2(\s,c))),
\label{eq:q(z|s,c)}
\\
\decdis(\s|\z,c) &=
\mathcal{N}(\s|\vmu_{\theta}(\z,c), {\rm diag}(\vsigma_{\theta}^2(\z,c))),
\label{eq:p(s|z,c)}
\end{align}
and the variational lower bound to be maximized becomes
\begin{align}
\mathcal{J}(\phi,\theta) 
=
&
\mathbb{E}_{
(\s,c)\sim p_{\rm D}(\s,c)
}\big[ 
\mathbb{E}_{\z\sim q(\z|\s,c)}[\log p(\s|\z,c)] 
\nonumber\\
&~~~~~~~~~~~~~~~~~~~~~~~~~~~~~~
- {\rm KL}[ q(\z|\s,c) \| p(\z) ]
\big],
\end{align}
where $\mathbb{E}_{(\s,c)\sim p_{\rm D}(\s,c)}[\cdot]$ denotes the sample mean over the training examples $\{\s_m,c_m\}_{m=1}^{M}$.

One notable feature as regards CVAEs is that 
they are able to learn a ``disentangled'' latent representation underlying the data of interest. 
For example, 
when a CVAE is trained using the MNIST dataset of handwritten digits
and $c$ as the digit class label, 
$\z$ and $c$ 
are disentangled so 
that $\z$ represents the factors of variation corresponding to handwriting styles. 
We can thus generate images of 
a desired digit with random handwriting styles
from the trained decoder 
by specifying $c$ and randomly sampling $\z$.
Analogously, we would be able to obtain
a generative model that can represent the spectrograms of a variety of sound sources
if we could train a CVAE using class-labeled training examples.

\subsection{Multichannel VAE}
\label{subsec:propalgo}

Let $\tilde{\S}=\{s(f,n)\}_{f,n}$ be the complex spectrogram of a particular sound source
and $c$ be the class label of that source. Here, 
we assume that a class label comprises
one or more categories, each consisting of multiple classes.
We thus represent $c$ as a concatenation of one-hot vectors,
each of which is filled with 1 at the index of a class in a certain
category and with 0 everywhere else. For example, if
we consider speaker identities as the only class category,
$c$ will be represented as a single one-hot vector, where each
element is associated with a different speaker.

We now model the generative model of $\tilde{\S}$ 
using a CVAE with an auxiliary input $c$. 
So that the decoder distribution 
has the same form as the LGM \refeq{LGM1}, 
we define it as a zero-mean complex
Gaussian distribution 
\begin{align}
\decdis(\tilde{\S}|\z,c,g) 
&=
\prod_{f,n} \mathcal{N}_{\mathbb{C}}(s(f,n)|0,v(f,n)),\\
v(f,n) &= g \cdot \sigma_{\theta}^2(f,n; \z,c),
\end{align}
where
$\sigma_{\theta}^2(f,n; \z,c)$ denotes the $(f,n)$-th element of the decoder output
$\vsigma_{\theta}^2(\z,c)$ 
and $g$ represents the global scale of the generated spectrogram.
As regards the encoder distribution $\encdis(\z|\tilde{\S},c)$, we 
adopt a regular Gaussian distribution
\begin{align}
\encdis(\z|\tilde{\S},c) = \prod_{k} \CN(z(k)|\mu_{\phi}(k;\tilde{\S},c),\sigma_{\phi}^2(k;\tilde{\S},c)),
\end{align}
where $z(k)$, $\mu_{\phi}(k;\tilde{\S},c)$ and $\sigma_{\phi}^2(k;\tilde{\S},c)$ represent the $k$-th elements of 
the latent space variable $\z$ and the encoder outputs 
$\vmu_{\phi}(\tilde{\S},c)$ and $\vsigma_{\phi}^2(\tilde{\S},c)$, respectively.
Given a set of labeled training examples $\{\tilde{\S}_m,c_m\}_{m=1}^{M}$,
we train the decoder and encoder NN parameters $\theta$ and $\phi$, respectively,
prior to source separation, 
using the training objective
\begin{align}
\mathcal{J}(\phi,\theta) 
=
&
\mathbb{E}_{
(\tilde{\S},c)\sim p_{\rm D}(\tilde{\S},c)
}\big[ 
\mathbb{E}_{\z\sim q(\z|\tilde{\S},c)}[\log p(\tilde{\S}|\z,c)] 
\nonumber\\
&~~~~~~~~~~~~~~~~~~~~~~~~~~~~~~
- {\rm KL}[ q(\z|\tilde{\S},c) \| p(\z) ]
\big],
\label{eq:VAE_training_objective}
\end{align}
where $\mathbb{E}_{(\tilde{\S},c)\sim p_{\rm D}(\tilde{\S},c)}[\cdot]$ denotes the sample mean over the training examples $\{\tilde{\S}_m,c_m\}_{m=1}^{M}$.
\reffig{cvae_training} shows the illustration of the present CVAE.

The trained decoder distribution $\decdis(\tilde{\S}|\z,c,g)$ can be used as 
a universal generative model that is able to generate 
spectrograms of all the sources involved in the training examples 
where 
the latent space variable $\z$, the auxiliary input $c$ and the global scale $g$ 
can be interpreted as the model parameters.
According to the properties of CVAEs,
we consider that the CVAE training promotes disentanglement between $\z$ and $c$ where 
$\z$ characterizes the factors of intra-class variation whereas 
$c$ characterizes the factors of categorical variation that represent source identities.
We call $\decdis(\tilde{\S}|\z,c,g)$ the CVAE source model. 

Since the CVAE source model is given in the same form as the LGM given by \refeq{LGM1},
we can develop a log-likelihood that has the same expression as \refeq{loglikelihood}
if we use $\decdis(\tilde{\S}_j|\z_j,c_j,g_j)$ to express the generative model of 
the complex spectrogram of source $j$.
Hence, we can search for a stationary point of the log-likelihood
by iteratively updating the separation matrices 
$\CW$, the global scale parameter $\CG=\{g_j\}_j$ and 
the VAE source model parameters $\Psi=\{\z_j,c_j\}_j$ 
so that the log-likelihood is guaranteed to be non-decreasing at each iteration.
We can use \refeqs{w_update1}{w_update2} to update $\CW$, 
backpropagation to update $\Psi$ and
\begin{align}
g_j \leftarrow \frac{1}{FN} \sum_{f,n} \frac{|y_j(f,n)|^2}{\sigma^2_{\theta}(f,n; \z_j,c_j)},
\label{eq:g_update}
\end{align}
to update $\CG$ where $y_j(f,n) = \wH_j(f)\x(f,n)$.
Note that \refeq{g_update} maximizes \refeq{loglikelihood} with respect to $g_j$
when $\CW$ and $\Psi$ are fixed.

The proposed algorithm is thus summarized as follows:
\begin{enumerate}
\itemsep=0mm
\item Train $\theta$ and $\phi$ using \refeq{VAE_training_objective}.
\item Initialize $\CW$, $\CG$ and $\Psi=\{\z_j,c_j\}_j$.
\item Iterate the following steps for each $j$:
\begin{enumerate}
\item Update $\w_j(0),\ldots,\w_j(F)$ using \refeqs{w_update1}{w_update2}.
\item Update $\psi_j=\{\z_j,c_j\}$ using backpropagation.
\item Update $g_j$ using \refeq{g_update}.
\end{enumerate}
\end{enumerate}

\begin{figure}[t!]
\centering
\begin{minipage}{.8\linewidth}
  \centerline{\includegraphics[width=.98\linewidth]{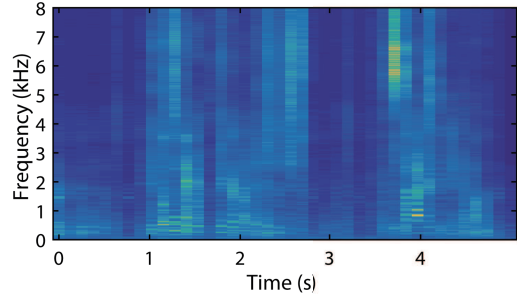}}
  \vspace{-1ex}
  \caption{Example of the CVAE source model fitted to the speech spectrogram shown in \reffig{nmfmodel}.}
\label{fig:cvae_source_model}
\end{minipage}
\end{figure}

The proposed MVAE is noteworthy in that it offers the advantages of the conventional methods concurrently. Namely, 
(1) it takes full advantage of the strong representation power of DNNs for source power spectrogram modeling,
(2) the convergence of the source separation algorithm is guaranteed, and
(3) the criteria for CVAE training and source separation are consistent, thanks to the consistency between 
the expressions of the CVAE source model and the LGM.
\reffig{cvae_source_model} shows an example of the CVAE source model fitted to the speech spectrogram shown in \reffig{nmfmodel}.
We can confirm from this example that the CVAE source model is able to approximate the speech spectrogram somewhat better than the NMF model.

\subsection{Network architectures}

\begin{figure*}[t!]
\centering
\centerline{\includegraphics[width=.65\linewidth]{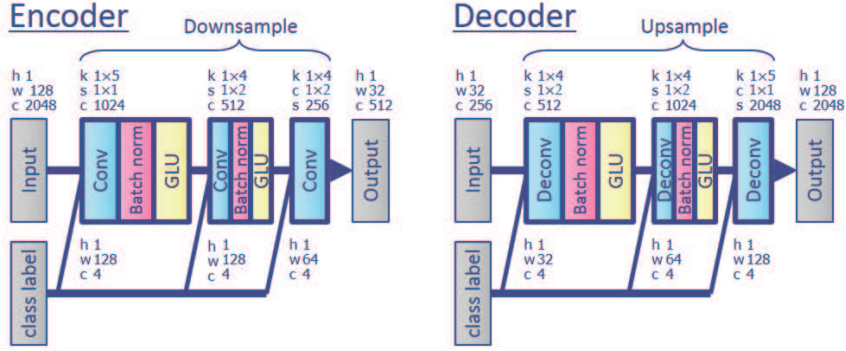}}
  \vspace{-0ex}
  \caption{Network architectures of the encoder and decoder. Here, the inputs and outputs of the encoder and decoder are interpreted as images, where ``h'', ``w'' and ``c'' denote the height, width and channel number, respectively. ``Conv'', ``Batch norm'', ``GLU'', ``Deconv'' denote convolution, batch normalization, gated linear unit, and transposed convolution layers, respectively. ``k'', ``s'' and ``c'' denote the kernel size, stride size and output channel number of a convolution layer, respectively. Note that all the networks are fully convolutional with no fully connected layers, thus allowing inputs to have arbitrary lengths.}
\label{fig:netarch}
\end{figure*}

We propose designing the encoder and decoder networks using fully convolutional architectures
to allow the encoder to take a spectrogram as an input and 
allow the decoder to output a spectrogram of the same length instead of a single-frame spectrum.
This allows the networks to capture time dependencies in spectral sequences.
While RNN-based architectures are a natural choice for modeling time series data, 
we use convolutional neural network (CNN)-based architectures to design the encoder and decoder as detailed below.

We use 1D CNNs to design the encoder and the decoder networks 
by treating $\tilde{\S}$ as an image of size $1 \times N$ with $F$ channels.
Specifically, we use a gated CNN \cite{Dauphin2017}, which 
was originally introduced to model word sequences for language modeling 
and was shown to outperform long short-term memory (LSTM) language models trained in a similar setting. 
We previously employed gated CNN architectures for voice conversion \cite{Kaneko2017c,Kaneko2017d,Kameoka2018} and monaural audio source separation \cite{Li2018}, and 
have already confirmed their effectiveness. 
In the encoder, the output of the
$l$-th hidden layer, $\sh_l$,   
is described as a linear projection 
modulated by an output gate
\begin{align}
\sh_{l-1}' &= [\sh_{l-1};\Vec{c}_{l-1}],
\label{eq:dec_glu_1}
\\
\sh_l &= (\sW_l * \sh_{l-1}' + \sb_l) \odot 
\sigma
(\sV_l * \sh_{l-1}' + \sd_l),
\label{eq:dec_glu_2}
\end{align}
where 
$\sW_l \in \mathbb{R}^{D_l\!\times\! D_{l-1}\!\times\! 1\!\times\! N_{l}}$, 
$\sb_l\in \mathbb{R}^{D_l}$, 
$\sV_l\in \mathbb{R}^{D_l\! \times\! D_{l-1}\!\times\! 1 \!\times\! N_{l}}$ and 
$\sd_l\in \mathbb{R}^{D_l}$ are
the encoder network parameters $\phi$, and 
$\sigma$ denotes the elementwise sigmoid function.
Similar to LSTMs,
the output gate multiplies each element of 
$\sW_l * \sh_{l-1} + \sb_l$
and controls what information should be propagated through the hierarchy of layers.
This gating mechanism is called a gated linear unit (GLU).
Here, $[\sh_{l};\Vec{c}_{l}]$ means the concatenation of 
$\sh_{l}$ and $\Vec{c}_{l}$
along the channel dimension, and
$\Vec{c}_l$ is 
a 2D array consisting of a $N_l$ tiling of copies of $c$ in the time dimensions.
The input into the 1st layer of the encoder is $\sh_0 = \tilde{\S}$.
The outputs of the final layer are 
given as regular linear projections
\begin{align}
\vmu_{\phi} &= \sW_L * \sh_{L-1}' + \sb_L,\\
\log \vsigma_{\phi}^2 &= \sV_L * \sh_{L-1}' + \sd_L.
\end{align}
The decoder network is devised in the same way as below with the only difference being
that $\vmu_{\theta}=\zero$:
\begin{align}
\sh_0 &= \z,\nonumber\\
\sh'_{l-1} &= [\sh_{l-1};\Vec{c}_{l-1}],
\nonumber\\
\sh'_l &= (\sW'_l * \sh'_{l-1} + \sb'_l) \odot 
\sigma
(\sV'_l * \sh'_{l-1} + \sd'_l),\nonumber\\
\vmu_{\theta} &= \Vec{0},\nonumber\\
\log \vsigma_{\theta}^2 &= \sV'_L * \sh'_{L-1} + \sd'_L,\nonumber
\end{align}
where $\sW'_l \in \mathbb{R}^{D_l\!\times\! D_{l-1}\!\times\!1\times\! N_l}$, 
$\sb'_l\in \mathbb{R}^{D_l}$, 
$\sV'_l\in \mathbb{R}^{D_l\! \times\! D_{l-1}\! \times\! 1\! \times N_l}$ 
and $\sd'_l\in \mathbb{R}^{D_l}$ are the decoder network parameters $\theta$.
It should be noted that the entire architecture is fully convolutional with no fully-connected layers.
The trained decoder can therefore be used a generative model of 
spectrograms with arbitrary lengths. This is particularly convenient 
when designing source separation systems since they can allow signals of any length. 

\section{Experiments}

\begin{figure}[t!]
\centering
\begin{minipage}{.65\linewidth}
  \centerline{\includegraphics[width=.98\linewidth]{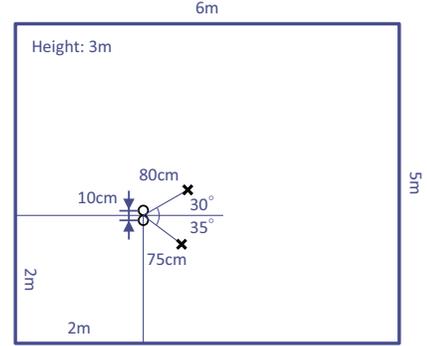}}
  \vspace{-1ex}
  \caption{Simulated room configuration.}
\label{fig:roomconfig}
\end{minipage}
\end{figure}

\begin{figure*}[t!]
\centering
\begin{minipage}{.33\linewidth}
  \centerline{\includegraphics[width=.95\linewidth]{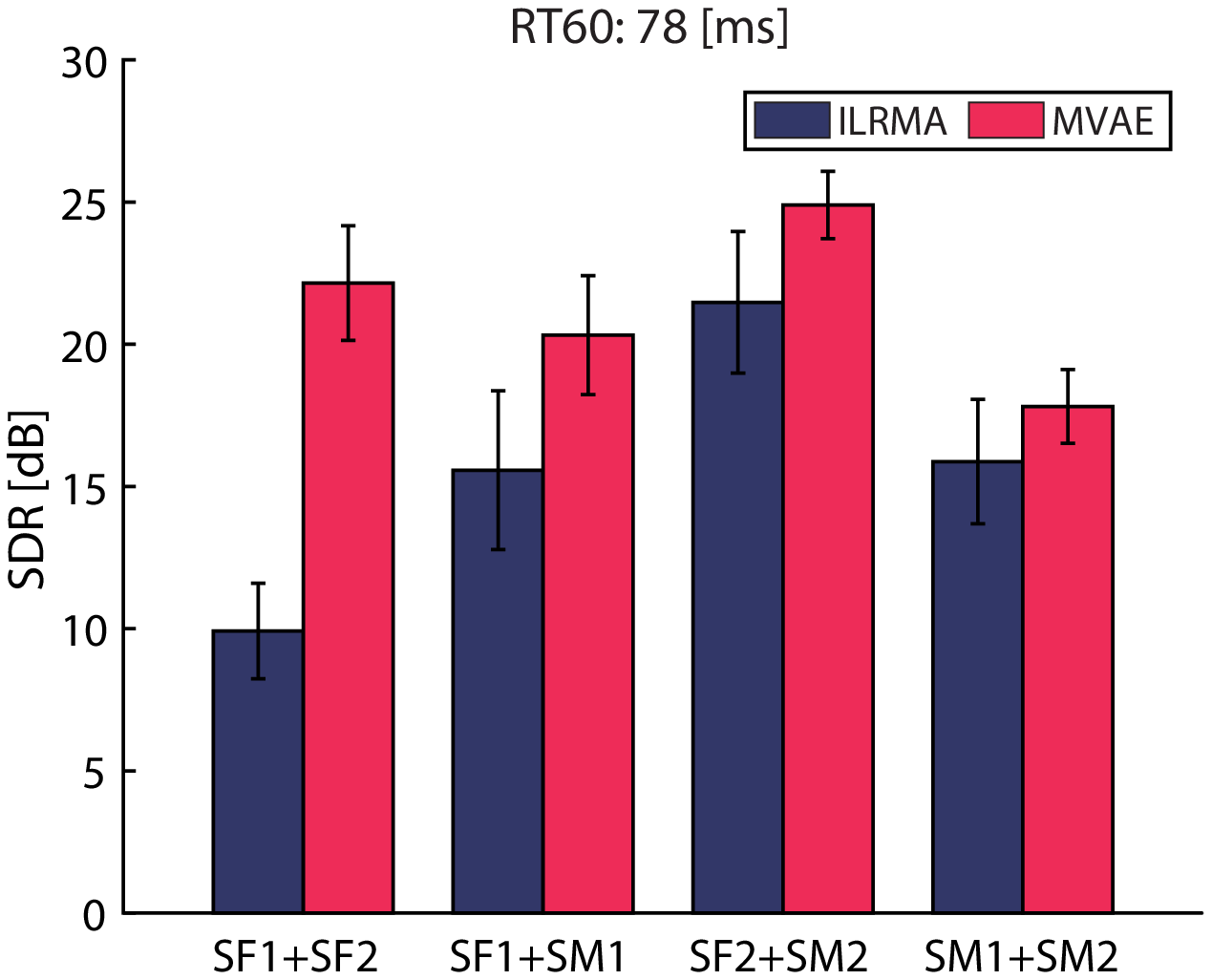}}
\end{minipage}
\begin{minipage}{.33\linewidth}
  \centerline{\includegraphics[width=.95\linewidth]{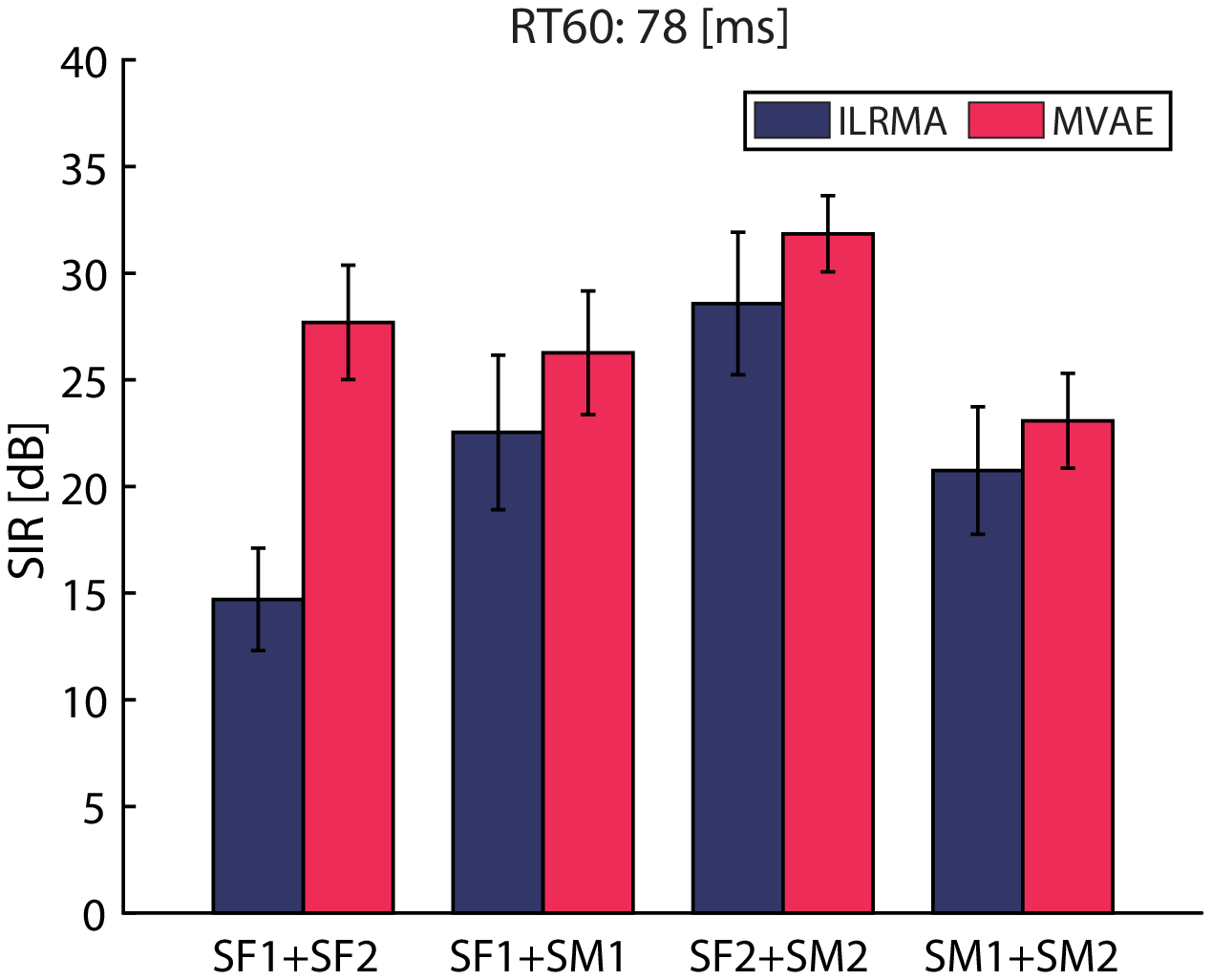}}
\end{minipage}
\begin{minipage}{.33\linewidth}
  \centerline{\includegraphics[width=.95\linewidth]{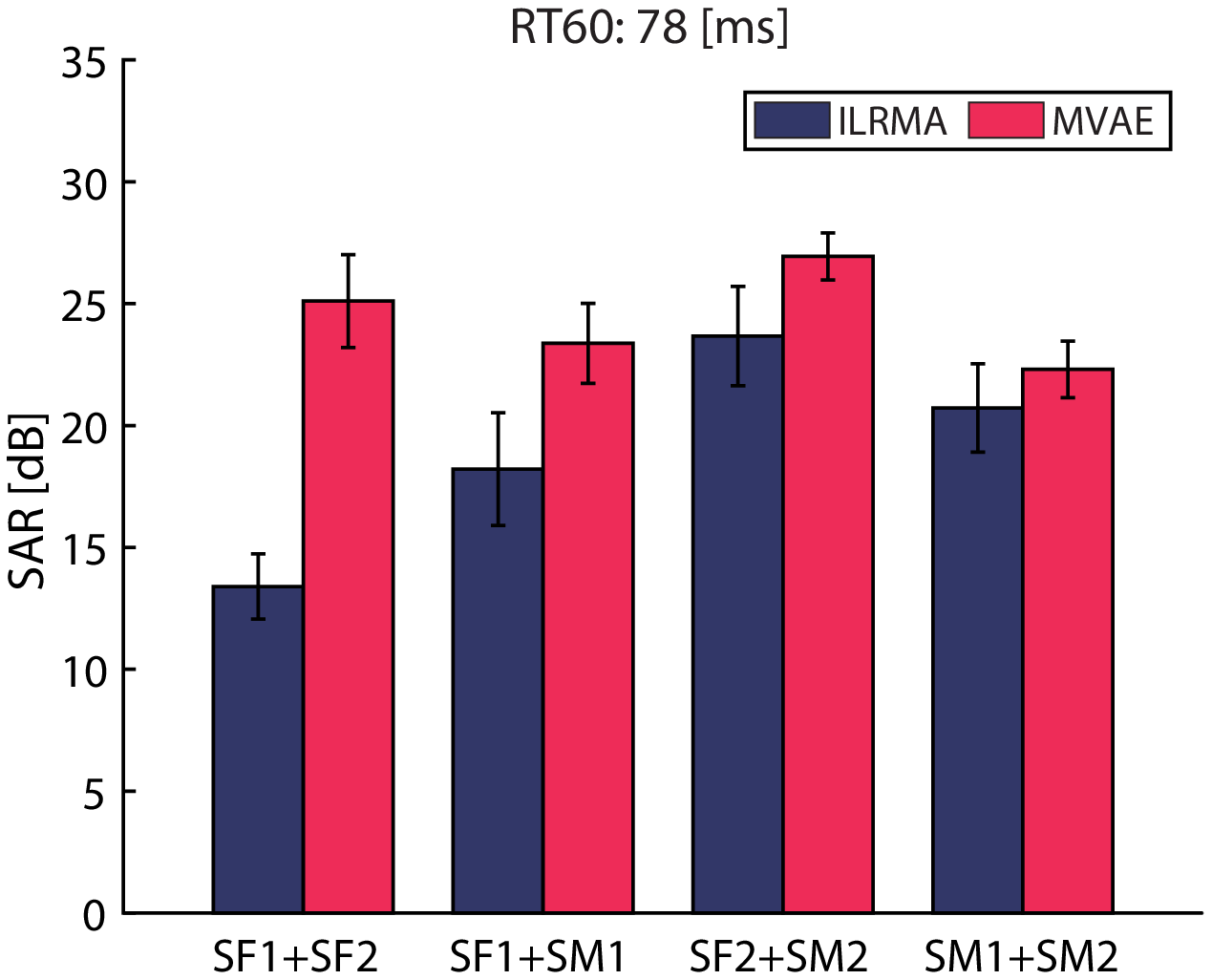}}
\end{minipage}
  \vspace{-1ex}
  \caption{Average SDRs, SIRs and SARs obtained with the baseline and proposed methods
  for ${\rm RT}_{60}$ of 78 [ms].}
\label{fig:SDR1}
\medskip
\begin{minipage}{.33\linewidth}
  \centerline{\includegraphics[width=.95\linewidth]{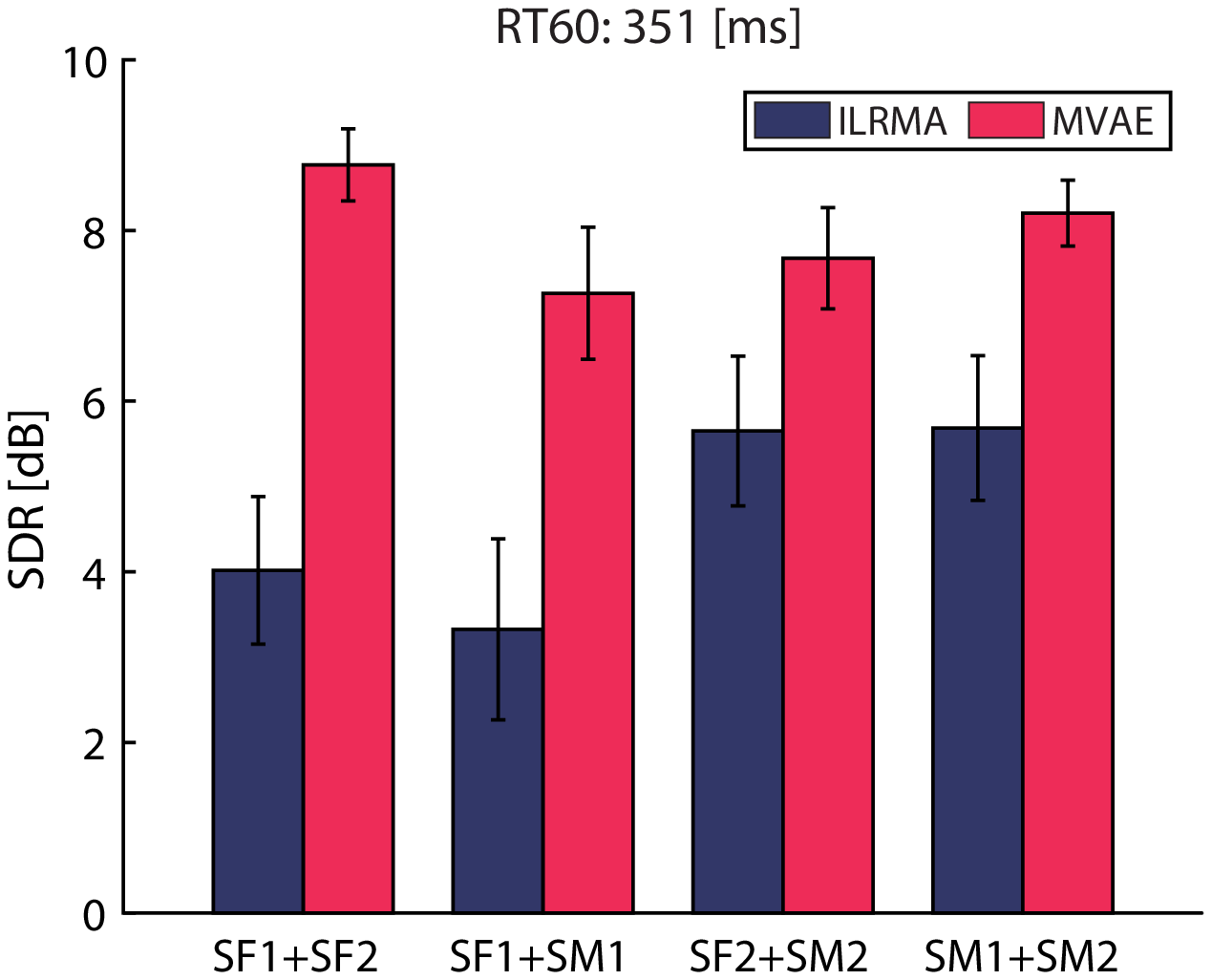}}
\end{minipage}
\begin{minipage}{.33\linewidth}
  \centerline{\includegraphics[width=.95\linewidth]{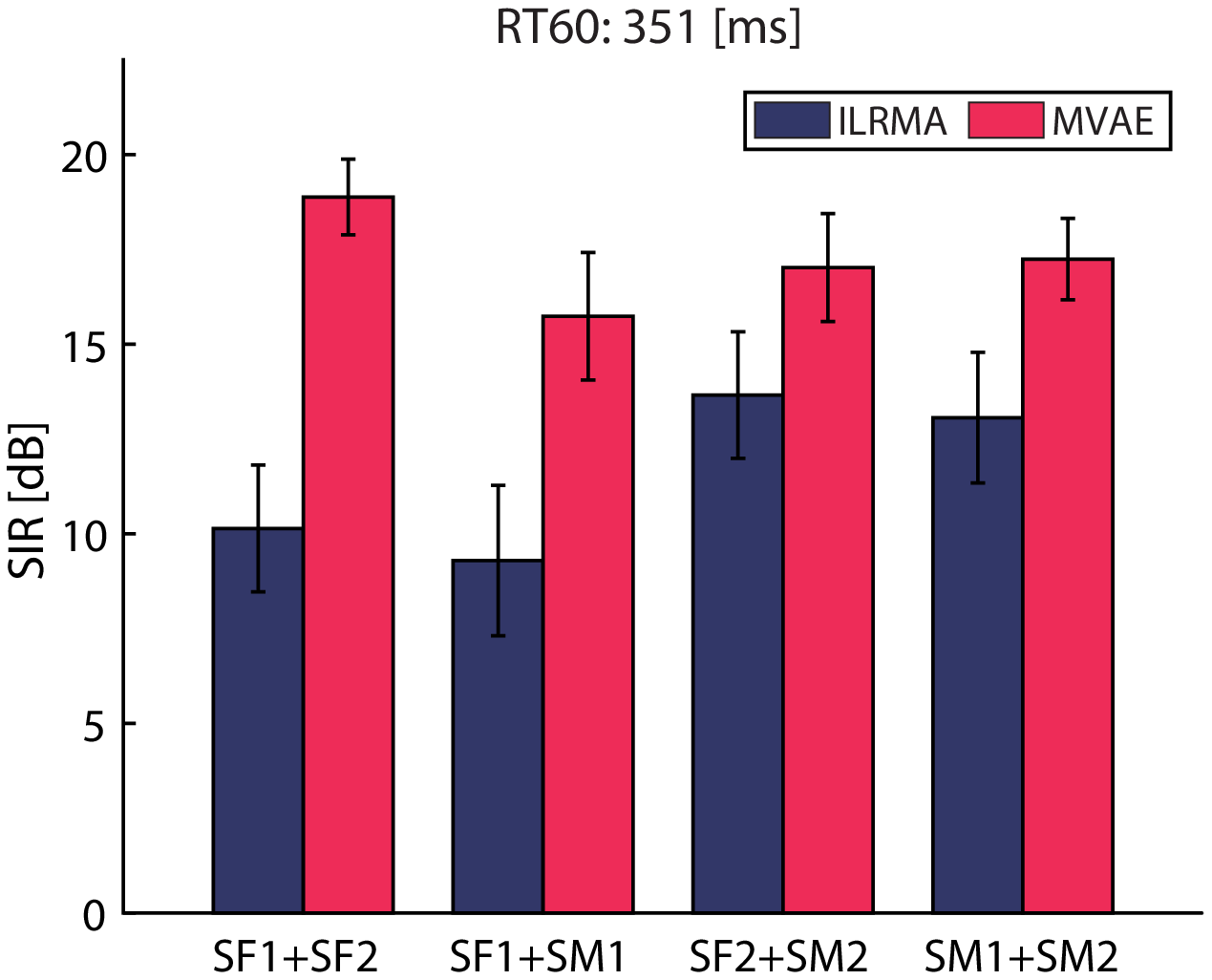}}
\end{minipage}
\begin{minipage}{.33\linewidth}
  \centerline{\includegraphics[width=.95\linewidth]{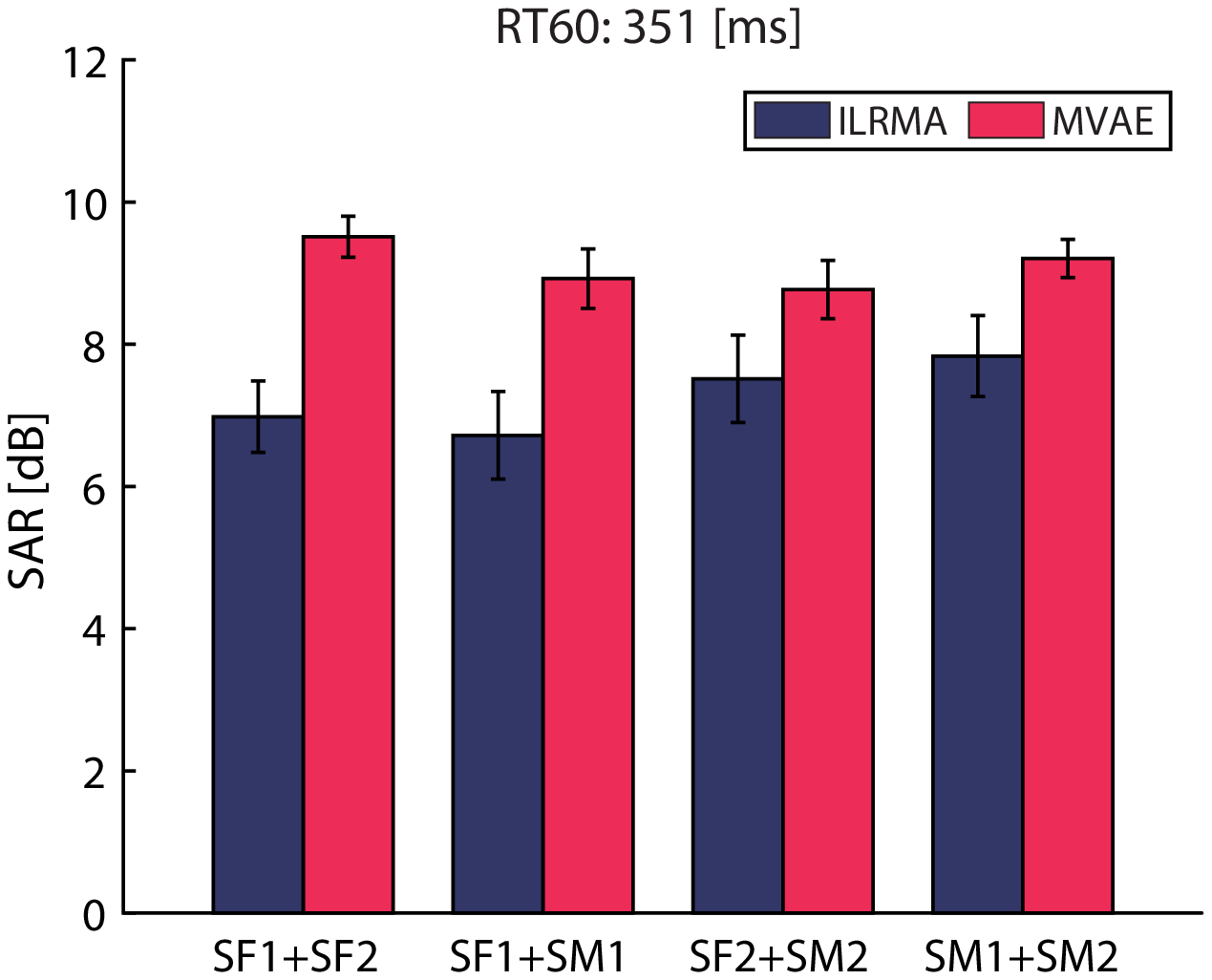}}
\end{minipage}
  \vspace{-1ex}
  \caption{Average SDRs, SIRs and SARs obtained with the baseline and proposed methods
  for ${\rm RT}_{60}$ of 351 [ms].}
\label{fig:SDR2}
\end{figure*}

To confirm the effect of the incorporation of the CVAE source model,
we conducted experiments involving a semi-blind source separation task using speech mixtures.
We excerpted speech utterances from 
the Voice Conversion Challenge (VCC) 2018 dataset \cite{Lorenzo-Trueba2018short},
which consists of recordings of six female and six male US English speakers. 
Specifically, we used the utterances of two female speakers, `SF1' and `SF2', 
and two male speakers, `SM1' and `SM2' for CVAE training and source separation. 
We considered speaker identities as the only source class category.
Thus, $c$ was a four-dimensional one-hot vector. 
The audio files for each speaker were manually
segmented into 116 short sentences (each about 7 minutes long)
where 81 and 35 sentences (about 5 and 2 minutes long, respectively) were provided
as training and evaluation sets, respectively. 
We used simulated two-channel recordings of two sources as the test data where 
the impulse responses were synthesized by using the image method.
\reffig{roomconfig} shows the two-dimensional configuration of the room. 
$\circ$ and $\times$ represent the positions of microphones and sources, respectively.
The reverberation time (${\rm RT}_{60}$) of the simulated signals could be controlled
according to the setting of the reflection coefficient of the walls. 
To simulate anechoic and echoic environments, we created test signals 
with the reflection coefficients set at 0.20 and 0.80, respectively. 
The corresponding ${\rm RT}_{60}$s were 78 [ms] and 351 [ms], respectively.
We generated 10 speech mixtures for each speaker pair, 
SF1+SF2, SF1+SM1, SM1+SM2, and SF2+SM2. 
Hence, there were 40 test signals in total, each of which was about 4 to 7 [s] long.
All the speech signals were re-sampled at 16000 [Hz].
The STFT frame length was set at 256 [ms] and a Hamming window was used with an overlap
length of 128 [ms]. 

We chose ILRMA \cite{Kameoka2010,Kitamura2016,Kitamura2017} as a baseline method for comparison.
The source separation algorithms were 
run for 40 iterations for the proposed method
and 100 iterations for the baseline method. 
For the proposed method, 
$\CW$ was initialized 
using the baseline method run for 30 iterations 
and
Adam optimization \cite{Kingma2015} was used for CVAE training and 
the estimation of $\Psi$ in the source separation algorithm.
The network configuration we used for the proposed method is shown in detail in \reffig{netarch}.
Note that we must take account of the sum-to-one constraints when 
updating $c_j$. This can be easily implemented by inserting an appropriately designed softmax layer
that outputs $c_j$
\begin{align}
c_j = {\rm softmax}(u_j),
\end{align}
and treating $u_j$ as the parameter to be estimated instead.

To evaluate the source separation performance,
we took the averages of the 
signal-to-distortion ratio (SDR), signal-to-interference ratio (SIR) and
signal-to-artifact ratio (SAR) \cite{Vincent2006} of the separated signals obtained with
the baseline and proposed methods
using 10 test signals for each speaker pair.
\reffigs{SDR1}{SDR2} show 
the average SDRs, SIRs and SARs obtained with the baseline and proposed methods
under different ${\rm RT}_{60}$ conditions. 
As the results show, the proposed method significantly outperformed the baseline method,
revealing the advantage of the CVAE source model.
Audio samples are provided at: 
http://www.kecl.ntt.co.jp/people/kameoka.hirokazu/ Demos/mvae-ass/.

As can be seen from a comparison between the results in \reffigs{SDR1}{SDR2},
there were noticeable performance degradations with both the baseline and proposed methods 
when the reverberation became relatively long. 
We hope that these degradations can be overcome by 
introducing the idea of jointly solving 
dereverberation and source separation, as in \cite{Kameoka2010,Yoshioka2011,Kagami2018}.

\section{Conclusions}

This paper proposed a multichannel source separation technique 
called the multichannel variational autoencoder (MVAE) method.
The method used VAEs to model and estimate the power spectrograms of the sources in mixture signals. 
The key features of the MVAE are that  
(1) it takes full advantage of the strong representation power of deep neural networks 
for source power spectrogram modeling,
(2) the convergence of the source separation algorithm is guaranteed, and
(3) the criteria for the VAE training and source separation are consistent,
which contributed to obtaining better separations than with conventional methods.


\small
\bibliographystyle{IEEEbib}
\bibliography{Kameoka2018arXiv07}

\end{document}